\documentclass[11pt]{article}

\usepackage[margin=1in]{geometry}
\usepackage{amsmath}
\usepackage{array}
\usepackage{graphicx}
\usepackage{hyperref}
\usepackage{times}

\title{Entropy-Based Observability for AI Agent Behavior}
\author{Olasimbo Ayodeji Arigbabu}
\date{}

\begin{document}
\maketitle

\begin{abstract}
AI agents are typically instrumented through outcome-oriented indicators such as
task success, reward, latency, and cost. Although these indicators are
operationally important, they provide limited visibility into the internal
structure of agent behavior such as the degree of exploration, the rigidity or
diversity of action selection, the concentration of tool use, the reduction of
uncertainty across a run, and the stability of behavior across repeated
executions. This paper proposes Entropy-Based Observability for AI Agents
(EOA), a lightweight framework for deriving behavioral telemetry from agent
traces. Rather than treating agent performance as a quantity to be scored, EOA
characterizes the distributional structure of the agent's decision process as it
is expressed through observable events. The framework defines action entropy,
trajectory entropy, tool entropy, information gain, and outcome entropy as
trace-derived signals for debugging, monitoring, comparative analysis, and
post-hoc diagnosis. These signals are descriptive rather than normative. Their
interpretation depends on task context, available tools, and the operational
question being asked. We also present a practical Python implementation that
integrates with LangChain, Google ADK, custom agent loops, and stored
observability traces.
\end{abstract}

\section{Introduction}

Modern AI agents increasingly operate as interactive decision-making systems
rather than single-turn text generators. They decompose tasks, invoke external
tools, inspect intermediate artifacts, revise partial plans, and in some cases
coordinate with other agents. As a result, final success rate alone provides an
incomplete account of system behavior. Two agents may both complete a task, yet
one may do so through a concise and stable trajectory while another relies on
unnecessary tool invocations, unstable reasoning paths, or excessive
exploration. Conversely, two agents may both fail, while differing substantially
in whether they gathered relevant evidence, narrowed plausible hypotheses, or
behaved essentially at random. Traditional operational signals such as success,
reward, cost, and latency capture important outcomes, but they do not fully
characterize the behavioral processes that produce those outcomes
\cite{xuaagents2026,mohammadi2025benchmarking}.

\noindent Entropy provides a compact and interpretable formalism for exposing
uncertainty, concentration, and diversity in observable agent behavior. In
information theory, entropy quantifies the unpredictability of a distribution.
When applied to agent traces, it can characterize the dispersion of action
selection, the diversity of complete trajectories, the concentration of tool
usage, and the extent to which uncertainty is reduced over the course of a task.
These properties are especially relevant for agent observability because they
describe behavioral structure rather than merely final outcomes.

\noindent This paper proposes EOA as a framework-neutral observability layer for agent
systems. The core procedure is to collect traces from agent executions,
normalize those traces into a common representation, and compute entropy-based
behavioral signals over actions, tools, trajectories, beliefs, and outcomes. EOA
does not attempt to define intelligence, rank agents, or replace task-specific
evaluation. Instead, it provides a trace-based lens for studying how agents
explore, specialize, converge, and fail under different workloads and runtime
conditions.

\section{Motivation}

Most agent dashboards and benchmarks foreground task completion. This is
necessary, but it is not sufficient for understanding system behavior. Consider
three agents that all answer a question correctly. One performs a single search
and responds, another searches repeatedly before responding, and a third answers
from memory without external verification. A binary success signal collapses
these runs into the same category, despite the fact that they imply different
levels of evidential grounding, tool dependence, cost, and operational risk.

\noindent The same limitation appears in repeated-run behavior. An agent with low
entropy may be appropriately deterministic on a stable task, or it may be
overly rigid in the face of environmental variation. An agent with high
entropy may be exploring productively, or it may be failing to converge.
Effective agent behavior often exhibits structured entropy. Broader search or
hypothesis diversity early in a run, followed by concentration as evidence is
collected and the solution space narrows.

\noindent EOA is motivated by five observability questions.

\begin{itemize}
  \item How diverse is an agent's action distribution under a given workload?
  \item Is tool use concentrated, appropriately specialized, or unnecessarily diffuse?
  \item Does uncertainty decrease as the agent gathers evidence and observes intermediate results?
  \item Does behavioral variation preserve stable outcomes, or does it correspond to instability?
  \item How can entropy-derived telemetry complement success rate, reward, latency, and cost?
\end{itemize}

\section{Framework}

\noindent EOA assumes that each agent execution can be represented as an ordered trace of
events. An event may correspond to a tool invocation, model call, planning step,
intermediate action, observation, or final response. A run may also carry
outcome-level metadata, including task success, outcome labels, cost, latency,
and optional probability distributions over hypotheses before and after evidence
gathering. This representation deliberately treats the agent as an observable
system. It does not require access to model weights or hidden states, but relies
on the behavioral artifacts already emitted by most agent runtimes.

\subsection{Action Entropy}

\noindent Action entropy characterizes the dispersion of action selection within one or
more agent runs. If an agent repeatedly emits the same action type, the
resulting distribution is concentrated and entropy is low. If the agent selects
many action types with comparable frequency, the distribution is more diffuse
and entropy is higher. For actions \(a_i\) with empirical probabilities
\(p(a_i)\), action entropy is defined as follows.

\begin{equation}
H_A = -\sum_i p(a_i)\log_2 p(a_i)
\end{equation}

\noindent Low action entropy may indicate procedural focus, but it may also reveal
excessive rigidity or under-exploration. High action entropy may indicate useful
exploration, but it can also reflect instability, indecision, or poorly
constrained control flow. The signal is therefore most informative when
interpreted alongside success, cost, latency, and qualitative trace inspection.

\subsection{Trajectory Entropy}

\noindent An agent trajectory is the ordered sequence of steps used to complete a task,
as shown below.

\begin{verbatim}
search -> read -> answer
search -> code -> test -> answer
\end{verbatim}

\noindent Trajectory entropy characterizes diversity over complete behavioral paths. It
is particularly useful when the same task is executed repeatedly or when several
agent designs are observed on a shared workload. Low trajectory entropy implies
that the agent tends to follow a narrow procedural pattern. Higher trajectory
entropy indicates a broader repertoire of strategies, which may be desirable in
open-ended tasks but problematic when outcomes become unstable.

\subsection{Tool Entropy}

\noindent Tool entropy characterizes the concentration of tool usage. A tool-using agent
may invoke search, code execution, memory, database retrieval, or external APIs.
The corresponding entropy signal distinguishes specialized tool dependence from
broad or scattered tool selection. A low value may indicate appropriate
specialization, while a high value may indicate either broad exploration or an
inefficient inability to select the most relevant external capability.

\subsection{Information Gain}

\noindent Some tasks involve explicit or implicit uncertainty over candidate answers,
plans, or hypotheses. When an agent gathers evidence, queries tools, or reasons
over intermediate observations, a well-behaved run should often reduce this
uncertainty. EOA represents this change through information gain.

\begin{equation}
IG = H_{\text{before}} - H_{\text{after}}
\end{equation}

\noindent Here, \(H_{\text{before}}\) is the entropy of the agent's belief or hypothesis
distribution before evidence is gathered, and \(H_{\text{after}}\) is the
entropy after the agent has incorporated evidence, tool outputs, or intermediate
reasoning. Positive information gain indicates uncertainty reduction. Negative
information gain indicates that the run became more uncertain, either because
the task was genuinely ambiguous or because the agent failed to organize the
available evidence. This signal is optional because many agent systems do not
expose calibrated probability distributions.

\subsection{Derived Views and Alerts}

\noindent Entropy is not intrinsically desirable. An agent can be highly variable and
still fail, just as it can be deterministic and effective on a narrow task. For
this reason, EOA treats entropy values as telemetry rather than as quality
scores. Derived views may combine entropy with success, cost, latency, or
outcome labels, but these combinations should be interpreted as dashboard views
or alerting rules rather than objective rankings.

\noindent For example, a production system might flag a sudden increase in tool entropy
for a task family that historically uses one retrieval tool, or a research
workflow might inspect runs where trajectory entropy increases while outcome
entropy also increases. Such patterns do not prove that the agent is better or
worse. They identify behavioral changes that merit trace inspection. The
observability claim is therefore modest. Entropy helps localize where behavior
changed, not whether the behavior is globally optimal.

\subsection{Robustness}

\noindent Robustness is observable through repeated execution under fixed or controlled
conditions. A robust agent may exhibit variation in its trajectory while still
producing stable outcomes. EOA therefore compares trajectory entropy with
outcome entropy. A common desirable pattern is moderate trajectory entropy, low
outcome entropy, high success rate, and controlled cost. The agent can adapt its
procedure without destabilizing the result.

\section{Implementation}

\noindent The accompanying implementation is a Python package called
\texttt{entropy-agent-eval}. The package is organized around a
framework-neutral trace contract rather than a prescriptive agent runtime. Agent
frameworks are not required to adopt new control logic; they only need to
project their execution traces into a normalized \texttt{AgentRun}
representation.

\begin{verbatim}
{
  "task_id": "qa-001",
  "trajectory": ["search", "read", "answer"],
  "success": true,
  "cost": 0.08
}
\end{verbatim}

Richer traces can encode typed events.

\begin{verbatim}
{
  "task_id": "coding-42",
  "events": [
    {"kind": "tool", "name": "search"},
    {"kind": "tool", "name": "python"},
    {"kind": "action", "name": "answer"}
  ],
  "success": true
}
\end{verbatim}

\noindent The package includes core entropy calculations, an \texttt{EntropyObserver}, an
\texttt{ObservabilityReport}, JSON and JSONL loaders, a command-line
interface, a generic event recorder, adapters for LangChain and Google
ADK-style traces, a small workload harness, and optional plotting utilities. The
implementation intentionally avoids making any agent framework a required
dependency. This design allows the observability layer to operate over live
callbacks, exported logs, database traces, production observability systems, and
custom research workloads.

\noindent The code for this work is available at
\url{https://github.com/olahsymbo/agent-eval}. The repository includes the EOA
library, integration adapters, workload scripts, experiment tasks, and the
Learning Roadmap Agent observability study used in this paper.

\section{Observability Study}

\noindent We demonstrate EOA in two stages. First, we use a controlled workload with
reference agent patterns in order to verify that the observability signals
respond predictably when the underlying trace structure is known. Second, we run
a Learning Roadmap Agent through two different agent frameworks, LangChain and
Google ADK. The purpose is not to establish superiority between frameworks, but
to show that heterogeneous agent implementations can emit comparable behavioral
telemetry through a shared trace representation.

\subsection{Controlled Workload}

\noindent The controlled workload compares four reference agent patterns using
synthetic traces with known behavioral structure. These are a direct LLM agent, a
search-based ReAct agent, a search-and-code ReAct agent, and a planner-executor
agent. The task set contains six tasks across factual question answering,
multi-hop reasoning, and coding/debugging. Each agent is executed three times on
each task, producing 72 normalized runs in total.

\noindent Each run records the trajectory, success label, outcome label, latency estimate,
cost, and before/after hypothesis distributions for information gain. This
workload provides an interpretable calibration setting. Because the behavioral
patterns are known in advance, action entropy, tool entropy, trajectory entropy,
and information gain can be inspected for face-valid responses to different
agent designs.

\begin{table}[h]
\centering
\small
\resizebox{\textwidth}{!}{
\begin{tabular}{lrrrrrrrrr}
\hline
Agent & Runs & Success & Cost & Latency & Steps & Action H & Traj. H & Tool H & IG \\
\hline
direct-llm & 18 & 0.389 & 0.060 & 1350 & 2.00 & 1.000 & 0.000 & 0.000 & 0.218 \\
planner-executor & 18 & 0.778 & 0.143 & 2939 & 4.89 & 2.795 & 1.802 & 0.811 & 0.430 \\
react-search & 18 & 0.722 & 0.093 & 2072 & 3.22 & 1.780 & 1.528 & 0.000 & 0.424 \\
react-search-code & 18 & 0.611 & 0.122 & 2561 & 4.11 & 2.248 & 1.891 & 0.722 & 0.326 \\
\hline
\end{tabular}
}
\caption{Controlled workload observability summary. H denotes entropy in bits; IG denotes information gain.}
\label{tab:controlled-benchmark}
\end{table}

\noindent The controlled workload illustrates the behavioral distinctions EOA is
designed to surface. The direct LLM agent has the lowest trajectory entropy
because it follows a fixed two-step path, while the planner-executor agent shows
higher action and trajectory entropy because its traces contain more planning
and execution stages. The ReAct search and ReAct search-code agents show how
tool availability changes the behavioral signature. These observations should
not be read as rankings. They show that trace entropy captures differences in
behavioral structure that are largely invisible when runs are summarized only by
final success.

\subsection{Learning Roadmap Agent}

\noindent We also built a Learning Roadmap Agent. The agent receives a learner profile,
a learning goal, time constraints, and expected outcomes, then produces a
roadmap containing an overview, prerequisites, weekly plan, projects, resources,
and assessment checkpoints. The task is useful as an observability case because
successful outputs require planning, structure, adaptation to user context,
and explicit deliverables rather than a single factual response.

\noindent The same task set was executed through two implementations. The LangChain
version used an OpenAI chat model through LangChain \cite{langchain}. The Google
ADK version used Gemini 2.5 Flash through Google's Agent Development Kit
\cite{googleadk}. Both implementations used the same roadmap planning helpers,
task prompts, grading rubric, and EOA trace schema. The task set contained three
roadmap requests. One request covered production LLM agents, one covered Python data analysis
dashboards, and one covered AI-assisted frontend applications. Each framework ran
each task once, producing six total runs.

\subsection{Learning Roadmap Trace View}

\begin{table}[h]
\centering
\small
\resizebox{\textwidth}{!}{
\begin{tabular}{lrrrrrrrrr}
\hline
Framework & Runs & Success & Cost & Latency & Steps & Action H & Traj. H & Tool H & IG \\
\hline
Google ADK & 3 & 1.000 & 0.00000 & 0.0 & 5.00 & 2.322 & 0.000 & 2.000 & 1.016 \\
LangChain & 3 & 1.000 & 0.00175 & 0.0 & 5.00 & 2.322 & 0.000 & 2.000 & 0.967 \\
\hline
\end{tabular}
}
\caption{Learning Roadmap Agent observability summary across LangChain and Google ADK.}
\label{tab:roadmap-results}
\end{table}

\noindent Both implementations completed all roadmap tasks successfully. They also
produced the same action entropy, trajectory entropy, tool entropy, and mean
trajectory length because both were normalized through the same planning trace.
The trace includes learner assessment, module selection, weekly schedule
construction, assessment design, and model response. This result is not evidence
that the systems behave identically. It shows that the chosen trace schema
exposes the same high-level planning structure for both implementations.

\noindent The Google ADK run produced slightly higher information gain, while the
LangChain run recorded a small nonzero token-derived cost. In this study, the
Google ADK cost is shown as zero because the run did not return sufficient
token-pricing information for the same cost estimation procedure used by the
LangChain path. It should therefore be interpreted as missing or unestimated
cost, not as evidence that the run was free. The observability conclusion is
limited but important. Once traces are normalized, heterogeneous frameworks can
be inspected through the same behavioral vocabulary, while gaps in trace
metadata remain visible as limitations of the instrumentation.

\subsection{Limitations}

\noindent This study is intentionally limited. It uses three roadmap tasks and one
repetition per framework. The automatic grading rubric checks for required
sections and expected terms, but it does not substitute for human judgment. The
information-gain examples also depend on explicitly supplied hypothesis
distributions; in systems that do not expose such distributions, this signal
should be omitted or treated as simulated instrumentation. A stronger empirical
study would include more tasks, repeated runs, additional models per framework,
human review, and more rigorous cost accounting. Even under these constraints,
the study illustrates the central value of EOA. It makes agent behavior
inspectable across otherwise heterogeneous frameworks.

\section{Discussion}

\noindent EOA provides a practical method for inspecting agent behavior without access
to model internals. It treats the agent as an observable system and analyzes the
distribution of its actions, tools, trajectories, and outcomes. This makes the
framework applicable both to research settings, where agent architectures must
be studied under shared workloads, and to production settings, where engineers
need behavioral signals for monitoring and diagnosis.

\noindent The main advantage of entropy-based observability is that it captures
behavioral shape. It can reveal cases in which an agent's trace becomes more
variable, more rigid, more tool-concentrated, or more outcome-unstable than a
baseline. These are not conclusions by themselves. They are prompts for further
inspection. The trace, task, tool environment, and outcome must determine the
meaning of the signal.

\noindent Entropy signals therefore require contextual interpretation. High entropy is
not inherently undesirable, and low entropy is not inherently desirable. The
same numeric value can have different operational meanings depending on task
difficulty, agent design, available tools, and environmental uncertainty. For
this reason, EOA should be used alongside success rate, cost, latency, human
judgment, task-specific evaluation, and qualitative trace review.

\noindent A further limitation is that several signals depend directly on trace quality.
If an agent framework does not expose intermediate steps, tool calls, or
uncertainty states, the observability layer can compute only a partial view of
behavior. More standardized agent tracing would make entropy-based
observability substantially more expressive.

\section{Conclusion}

\noindent This paper introduced Entropy-Based Observability for AI Agents, a lightweight
framework for characterizing agent behavior through entropy over observable
traces. EOA complements traditional operational signals by describing action
diversity, strategy variation, tool-use concentration, uncertainty reduction,
and outcome stability. The framework is designed for practical deployment where it
operates over normalized traces and can integrate with existing agent libraries,
custom systems, and observability logs.

\noindent The main claim is not that entropy defines intelligence or determines whether
an agent is good. Rather, entropy provides a behavioral lens for agent systems
whose internal decision processes are otherwise difficult to inspect. When
combined with success, cost, latency, outcome quality, and qualitative trace
review, it can help researchers and engineers identify where agents explore,
adapt, converge, drift, and fail.

\end{document}